\documentclass{article}

\PassOptionsToPackage{numbers, compress}{natbib}

\usepackage[preprint]{neurips_2024}

\usepackage[utf8]{inputenc} 
\usepackage[T1]{fontenc}    
\usepackage{hyperref}       
\usepackage{url}            
\usepackage{booktabs}       
\usepackage{amsfonts}       
\usepackage{nicefrac}       
\usepackage{microtype}      
\usepackage{xcolor}         

\usepackage{enumitem}
\usepackage{graphicx}

\usepackage{amsmath}

\title{Higher-order-ReLU-KANs (HRKANs) for\\ solving physics-informed neural networks (PINNs)\\ more accurately, robustly and faster}

\author{%
  Chi Chiu SO \\
  College of Professional and Continuing Education\\
  The Hong Kong Polytechnic University\\
  Hong Kong, China\\
  \texttt{kelvin.so@cpce-polyu.edu.hk} \\
  \And 
  Siu Pang YUNG \\
  Department of Mathematics\\
  The University of Hong Kong \\
  Hong Kong, China \\
  \texttt{spyung@hku.hk} \\
}

\begin{document}

\maketitle

\begin{abstract}
    Finding solutions to partial differential equations (PDEs) is an important and essential component in many scientific and engineering discoveries. One of the common approaches empowered by deep learning is Physics-informed Neural Networks (PINNs). Recently, a new type of fundamental neural network model, Kolmogorov-Arnold Networks (KANs), has been proposed as a substitute of Multilayer Perceptions (MLPs), and possesses trainable activation functions. To enhance KANs in fitting accuracy, a modification of KANs, so called ReLU-KANs, using "square of ReLU" as the basis of its activation functions, has been suggested. In this work, we propose another basis of activation functions, namely, Higher-order-ReLU (HR), which 
    \begin{itemize}
        \item is simpler than the basis of activation functions used in KANs, namely, B-splines;
        \item allows efficient KAN matrix operations; and
        \item possesses smooth and non-zero higher-order derivatives, essential to physics-informed neural networks.
    \end{itemize}
    We name such KANs with Higher-order-ReLU (HR) as their activations, HRKANs. Our detailed experiments on two famous and representative PDEs, namely, the linear Poisson equation and nonlinear Burgers' equation with viscosity, reveal that our proposed Higher-order-ReLU-KANs (HRKANs) achieve the highest fitting accuracy and training robustness and lowest training time significantly among KANs, ReLU-KANs and HRKANs. The codes to replicate our experiments are available at \url{https://github.com/kelvinhkcs/HRKAN}.
\end{abstract}

\section{Introduction}

Partial differential equations (PDEs) play a prominent role in many real-world scenarios for modelling the dynamics in the systems of interest. 
Solving PDEs is essential for us to learn the dynamics of many scientific and engineering phenomena. However, it is often difficult, if not impossible, to find an exact solution of PDEs, especially when the PDE systems or the boundary or initial conditions are too complex~\cite{larsson2003partial, hughes2005isogeometric}. A variety of numerical methods thus become desirable given their ease of use and wide applicability in finding approximate solutions for PDEs. Common examples include finite element methods~\cite{hughes2012finite}, mesh-free methods~\cite{rabczuk2019extended}, finite difference methods~\cite{leveque2007finite}, finite volume methods~\cite{darwish2016finite}, and boundary element methods~\cite{brebbia2012boundary}.

"AI for Science" is a recently emerging field of AI, which comprises of inventing and using deep learning algorithms to solve PDEs. Among these algorithms, there are four major approaches:
\begin{enumerate}
    \item Physics-informed Neural Networks (PINNs)~\cite{raissi2019physics}: In this approach, neural networks are designed to approximate the solution of PDEs. The loss functions of neural networks are defined based on the PDE equations, with the aim to train the neural networks to become the solutions of the PDEs. PINNs have been widely adopted in solving PDEs~\cite{lu2021deepxde}, identifying PDEs from data~\cite{haghighat2020deep}, and obtaining optimal controls of PDEs~\cite{cao2024system, mowlavi2023optimal}. 
    
    \item Operator learning: In this approach, neural networks are designed to learn the operator mappings from PDE space to the solution space from a massive data-set~\cite{lu2022comprehensive}, key representatives of which include DeepONet~\cite{lu2021learning, lu2019deeponet} and Fourier Neural Operator (FNO)~\cite{li2020fourier}. A key difference between PINNs and operator learning is that when the initial or boundary conditions change, PINNs need to be re-trained with a new loss function or additional data points, whereas the operators do not require any re-training. Instead, the operators can immediately provide solutions even when those conditions change~\cite{kovachki2023neural}.
    
    \item Physics-informed Neural Operators~\cite{wang2021learning, li2024physics, goswami2023physics}: This approach combines PINNs with operator learning such that the operators learn the mappings from the PDE space to the solution space based on a physics-informed loss function, without the necessary need of big data-set.
    
    \item Neural Spatial Representations and PDE-Nets: These models are proposed to solve time-dependent PDEs. Neural Spatial Representations~\cite{chen2023implicit} use neural networks to represent spatially-dependent functions to map the solutions at a time instance to the next time instance. PDE-Nets~\cite{long2018pde, so2021differential} design tailor-made convolution filters to act as differential operators and apply forward Euler method to map the solutions at a time instance to the next time instance.
\end{enumerate}

Underlying all of these four approaches, the backbone of the neural network models is Multilayer perceptrons (MLPs)~\cite{haykin1998neural,cybenko1989approximation}, also known as fully-connected feedforward neural networks. The famous universal approximation power possessed by MLPs~\cite{hornik1989multilayer} makes it the most essential building block in deep learning applications, ranging from self-driving cars~\cite{bachute2021autonomous} to textual and video generations~\cite{wu2023brief}.

Recently, a new architecture of neural network building block, Kolmogorov-Arnold Networks (KANs)~\cite{liu2024kan}, has been proposed as a substitute of MLPs. Different from MLPs, KANs are not backed by the universal approximation theorem. Instead, KANs are supported by the Kolmogorov-Arnold representation theorem~\cite{kolmogorov1961representation}, which states that any multivariate function can be expressed by a finite combination of univariate functions~\cite{braun2009constructive}. Another key difference between MLPs and KANs lie on their parameter spaces. The parameter space of MLPs solely consists of weight matrices, whereas the parameter space of KANs also includes the activation functions. In other words, KANs need to learn not only the optimal weight matrices, but also the optimal activation functions. In the original invention of KANs~\cite{liu2024kan}, B-splines are used as the basis of activation functions to be learnt. Afterwards, suggestions of different bases of activation functions have been seen, including Chebyshev orthogonal polynomials~\cite{ss2024chebyshev}, radial basis functions~\cite{li2024kolmogorov}, wavelet transforms~\cite{bozorgasl2024wav}, Jacobi basis functions~\cite{aghaei2024fkan, aghaei2024rkan}, Fourier transform~\cite{xu2024fourierkan} and "square of ReLU" basis functions~\cite{qiu2024relu}.

Some efforts have also been witnessed in trying to apply KANs to find solutions of PDEs, for example, by incorporating KANs in DeepONets~\cite{abueidda2024deepokan} and putting KANs in PINNs~\cite{wang2024kolmogorov, wu2024ropinn, howard2024finite, calafa2024physics}. However, due to the complexity of KANs' B-spline basis functions, the training speed of KANs is not comparable to MLPs. One possible remedy is the use of "square of ReLU" basis functions in ReLU-KANs, which optimize KAN operations for efficient GPU parallel computing. A significant drawback, nevertheless, is the discontinuity of higher-order derivatives of such "square of ReLU" basis functions.

In this paper, we introduce a more suitable basis function, Higher-order-ReLU (HR), which not only optimizes KAN operations for efficient GPU parallel computing but at the same time is well-suited for PINNs. We call such KANs using Higher-order-ReLU basis functions (HR) as HRKANs, and such PINNs using HRKANs as PI-HRKANs. We evaulated PI-HRKANs' performances on two representative and famous PDEs, namely, the linear Poisson equation (the example used in the github codebase of the original KAN paper), and the nonlinear Burgers' equation with viscosity. Compared against KANs and ReLU-KANs, HRKANs exhibit significant improvements in fitting accuracy, training robustness and convergence speed. The codes to replicate our experiments are available at \url{https://github.com/kelvinhkcs/HRKAN}.

This paper presents the following key contributions:
\begin{itemize}
    \item \textbf{A simple activation basis function}: We introduce a simple activation basis function, Higher-order-ReLU (HR), which inherits the fitting capability of the original KAN B-spline basis functions and the "square of ReLU" basis functions.
    \item \textbf{Matrix-based operations}: Similar to the "square of ReLU" basis functions in ReLU-KAN, Higher-order-ReLU (HR) basis functions facilitate efficient matrix computations with GPUs, speeding up the training process.
    \item \textbf{Well-suited for Physics-informed problems}: While enabling efficient matrix operations and being simpler than the B-spline basis functions, Higher-order-ReLU (HR) basis functions also possess smooth and non-zero higher-order derivatives (with order up to users' choices), thus being suitable for Physics-informed problems. Such smooth and non-zero higher-order derivatives are missing in "square of ReLU" basis functions as we shall show in later sections.
    \item \textbf{Higher fitting accuracy, stronger robustness and faster convergence}: HRKANs demonstrate highest fitting accuracy, strongest robustness and fastest convergence compared to KANs and ReLU-KANs in our detailed experiments.
\end{itemize}

In section~\ref{s2}, we introduce the B-spline activation basis functions in KANs and the "square of ReLU" activation basis functions in ReLU-KANs and their respective properties. In section~\ref{s3}, we introduce our proposed Higher-order-ReLU (HR) activation basis functions, and discuss how it is better than the B-splines and "square of ReLU". In section~\ref{s4}, we conduct detailed and comprehensive experiments to evaluate HRKANs against KANs and ReLU-KANs, in terms of convergence speed, robustness and fitting accuracy. Section~\ref{s5} is our conclusion.

\section{Basis of activation function in KANs and ReLU-KANs} \label{s2}

In this section, we review the bases of activation functions used in KANs and ReLU-KANs, along with some comparisons. 

\subsection{KANs: B-spline basis function}

In KANs, B-splines are used as the basis of activation functions. We use $B_{g,k} = \{b_0(x), b_2(x), \ldots, b_n-1(x)\}$ to denote a set consisting of $n=g+k$ B-spline basis functions with $g$ grid points and order $k$. Each $b_i$ is defined recursively from linear combinations of B-spline basis functions of lower orders, i.e. order less than $k$~\cite{prautzsch2002bezier}. B-Spline basis functions are of compact support and their shapes depend on $k$ and $g$. 

\begin{figure}
  \makebox[\textwidth][c]{\includegraphics[width=1\textwidth]{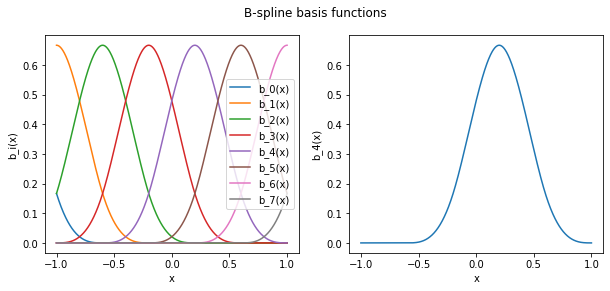}}%
\caption{B-Spline basis $B_{5,3}$ (left) and basis function $b_4(x)$ (right)}\label{bspline}
\end{figure}

In figure~\ref{bspline}, the left hand side shows the B-spline basis $B_{5,3} = \{b_0(x), b_1(x), \ldots, b_7(x)\}$ on the interval $[-1,1]$ with $g=5$ and $k=3$, and the right hand side shows the $b_4$ in the B-spline basis on the interval $[-1,1]$ with $g=5$ and $k=3$.

\subsection{ReLU-KANs: "square of ReLU" basis function}

In ReLU-KANs, the $b_i(x)$ with support on $[s_i,e_i]$ is replaced by the "square of ReLU" basis $r_i(x)$ defined by
\begin{equation}
    r_i(x) = [\text{ReLU}(e_i-x) \times \text{ReLU}(x-s_i)]^2 \times c, 
\end{equation}
where $c = \frac{16}{(e_i-s_i)^4}$ is a normalization constant for ensuring the height of each basis being 1. The set containing the $n=g+k$ "square of ReLU" basis functions is denoted $R_{5,3}$.

\begin{figure}
  \makebox[\textwidth][c]{\includegraphics[width=1\textwidth]{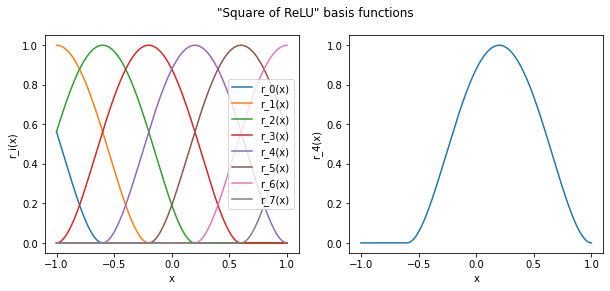}}%
\caption{"Square of ReLU" basis $R_{5,3}$ (left) and basis function $r_4(x)$ (right)}\label{sqrelu}
\end{figure}

Similar to the B-spline basis functions, the $r_i(x)$ is of compact support. Their shapes depend on the choice of $e_i$ and $s_i$. Figure~\ref{sqrelu} shows the "square of ReLU" basis functions and $r_4(x)$ on the interval $[-1,1]$ defined with the same support as the B-splines basis function example in figure~\ref{bspline}.

\section{Our method: Higher-order-ReLU} \label{s3}

In this section, we introduce our proposed basis function, namely, Higher-order-ReLU (HR), an improvement from the "square of ReLU". Higher-order-ReLU looks similar to "square of ReLU", but is better than "square of ReLU" in the following ways:
\begin{enumerate}
\item Higher-order-ReLU offers smooth and non-zero higher-order derivatives, essential to Physics-informed problems, whereas "square of ReLU" has discontinuous derivatives.
\item Higher-order-ReLU still inherits all advantages of “square of ReLU" on its simplicity and efficient matrix-based KAN operations over the B-splines.
\end{enumerate}

The Higher-order-ReLU basis $v_i$ of order $m$ is defined as
\begin{equation}
    v_{m,i}(x) = \text{ReLU}(e_i-x)^m \times \text{ReLU}(x-s_i)^m \times c_m
\end{equation}
where $c_m = \left(\frac{2}{(e_i-s_i)}\right)^{2m}$ is the normalization constant for ensuring the height of each basis being 1.

\begin{figure}
  \makebox[\textwidth][c]{\includegraphics[width=1\textwidth]{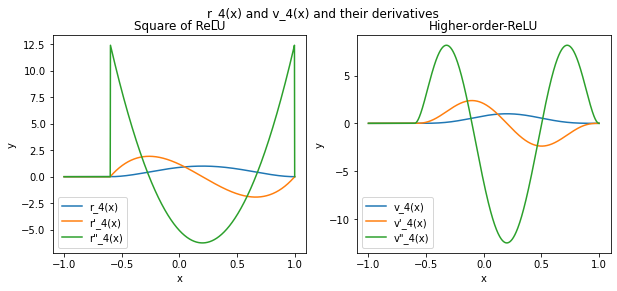}}%
\caption{The "square of ReLU" $r_4(x)$ and its first and second-order derivatives $r'_4(x)$ and $r''_4(x)$ (left) and the Higher-order-ReLU $v_{4,4}(x)$ and its first and second-order derivatives $v'_{4,4}(x)$ and $v''_{4,4}(x)$ (right).}\label{higher_order}
\end{figure}

A key difference between the $r_i(x)$ in "square of ReLU" and the $v_{m,i}(x)$ in Higher-order-ReLU lies on the order $m$ of $v_{m,i}(x)$, which determines the smoothness of higher-order derivatives of Higher-order-ReLU. 
Figure~\ref{higher_order} compares the Higher-order-ReLU $v_{m,i}(x)$ with $m=4$ and "square of ReLU" $r_i(x)$ and their first and second-order derivatives. It is obvious that the second-order derivative of "square of ReLU" jumps at the end points of its support, which hinders the learning of PDE solution. If a PDE equation possesses derivative terms of order higher than 4, for example, $u_{xxxxx}$, we can pick $m$ to be larger, like 6, in an adaptive manner, whereas its "square of ReLU" basis $R_{5,3}$ counterpart will fail as its 5-th order derivatives are completely zero. 

Comparing with the B-splines in KANs, we can observe that the higher-order derivatives of the B-spline basis are smooth, but they vanish up to the order $k$ specified. So, again, if the PDE equation possesses derivative terms of order higher than 4, for example, $u_{xxxxx}$, then the $k$ must be set to be higher than 6 to work out successfully. In this case, as B-splines are essentially polynomial, having an order which is too high may lead to the Runge's phenomenon~\cite{epperson1987runge}, an obstacle in function approximation which is not negligible.

\section{Experiments} \label{s4}

In this section, we evaluate the performance of KANs, ReLU-KANs and HRKANs (with $m=4$) with two physics-informed problems. The first one is a linear PDE, the Poisson equation. The second one is a non-linear PDE, the Burgers' equation with viscosity. 

\subsection{Poisson Equation}

\paragraph{Problem Description.}

The 2D Poisson equation is 
\begin{equation}
    \nabla^2 u(x,y) = -2\pi^2 \sin(\pi x)\sin(\pi y)
\end{equation}
for $(x,y) \in \left[-1,1\right] \times \left[-1,1\right]$,
with boundary conditions
\begin{equation*}
  \left\{
    \begin{aligned}
        &u(-1,y) = 0\\
        &u(1,y) = 0\\
        &u(x,-1) = 0\\
        &u(x,1) = 0
    \end{aligned}.
  \right.
\end{equation*}

The ground-truth solution is $$u(x,y) = \sin(\pi x)\sin(\pi y).$$ In the train-set, the number of interior points are $N_{\text{pde}}= 960$, and number of boundary points on the 4 boundaries the same, namely, $N_{\text{bc1}} = 30, N_{\text{bc2}} = 30, N_{\text{bc3}} = 30$ and $N_{\text{bc4}} = 30$ respectively. The train-set data points on the interior are generated randomly with uniform distribution while those on the boundaries are generated uniformly. We generate test-set data with a grid of grid size $100 \times 100$.

We trained a KAN, a ReLU-KAN and a HRKAN using the loss function
\begin{equation}
L = \alpha L_{\text{pde}} + L_{\text{bc}}
\end{equation}
where \begin{align*}
    L_{\text{pde}} &= \frac{1}{N_{\text{pde}}}\sum_{i=1}^{N_{\text{pde}}} \left( \nabla^2 u(x_i,y_i) + 2\pi^2 \sin(\pi x_i)\sin(\pi y_i) \right)^2,\\
    L_{\text{bc}} &= \frac{\sum_{i=1}^{N_{\text{bc1}}} \left(u(-1,y_i)\right)^2 + \sum_{i=1}^{N_{\text{bc2}}} \left(u(1,y_i)\right)^2 + \sum_{i=1}^{N_{\text{bc3}}} \left(u(x_i,-1)\right)^2 + \sum_{i=1}^{N_{\text{bc4}}} \left(u(x_i,1)\right)^2}{N_{\text{bc1}} + N_{\text{bc2}} + N_{\text{bc3}} + N_{\text{bc4}}}.
\end{align*}
We set hyperparameter $\alpha = 0.05$. The structure of the KAN, ReLU-KAN and HRKAN are all of $[2,2,1]$ with same support for all the bases with $g=5$ and $k=3$. For all three of them, we use the Adam optimizer, set the number of epoch to 3000 and train on the same GPU. We run the same experiment for 10 times. 

Two remarks have to be made here. 
\begin{enumerate}
    \item In the default implementation of KAN, there is a call of the function \texttt{update\_grid\_from\_samples()} every five epoch till epoch 50. We keep this implementation unchanged for KANs in both experiments of Poisson equation and Burgers' equation with viscosity, so KAN can update its grid whereas ReLU-KANs and HRKANs cannot in our experiments.
    \item In the default implementation of KAN, there is a also a step of sparsification for pruncing. Again, we have not included this implementation in ReLU-KANs and HRKANs but keep the implementation in KANs unchanged. 
\end{enumerate}
Both of these features may lead to extra overhead in computation, but we aim to keep the code of KAN unchanged to be fair. We will discuss more about our consideration in the section of result discussion.

\paragraph{Results Discussion.}

\begin{figure}
  \makebox[\textwidth][c]{\includegraphics[width=1.8\textwidth]{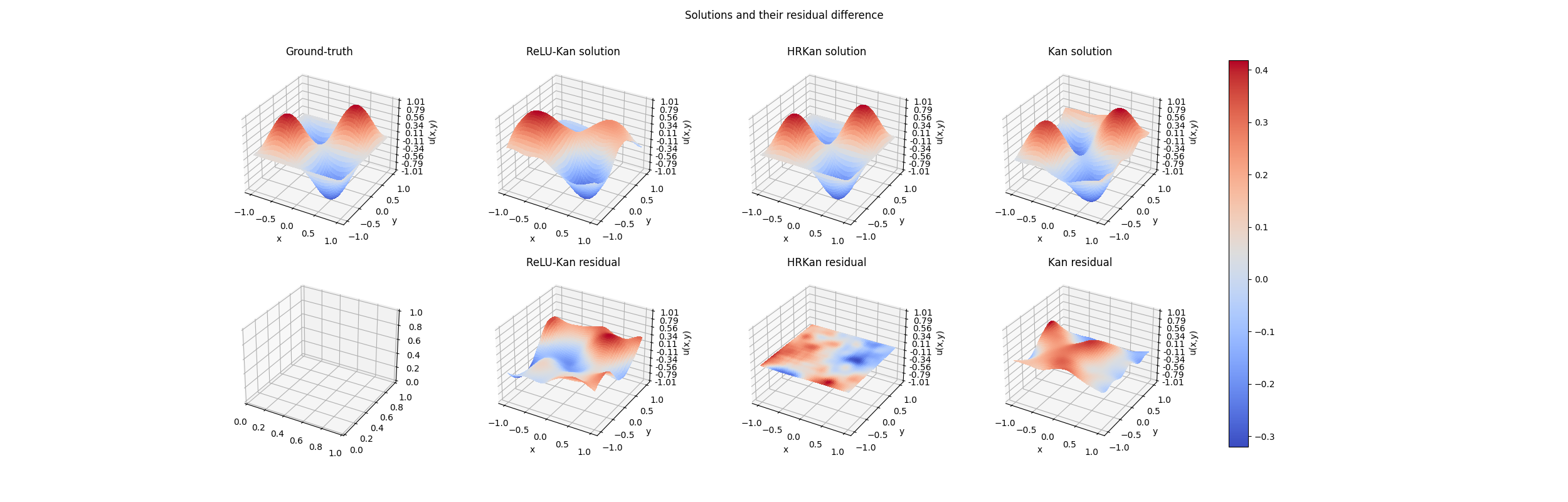}}%
\caption{The ground-truth solution and the solutions learnt by the KAN, ReLU-KAN and HRKAN (Top row) and their residual difference compared against the ground-truth solution (Bottom row) in one of the 10 runs with the Poisson equation.}\label{poisson_fig}
\end{figure}

\begin{figure}
  \makebox[\textwidth][c]{\includegraphics[width=1.8\textwidth]{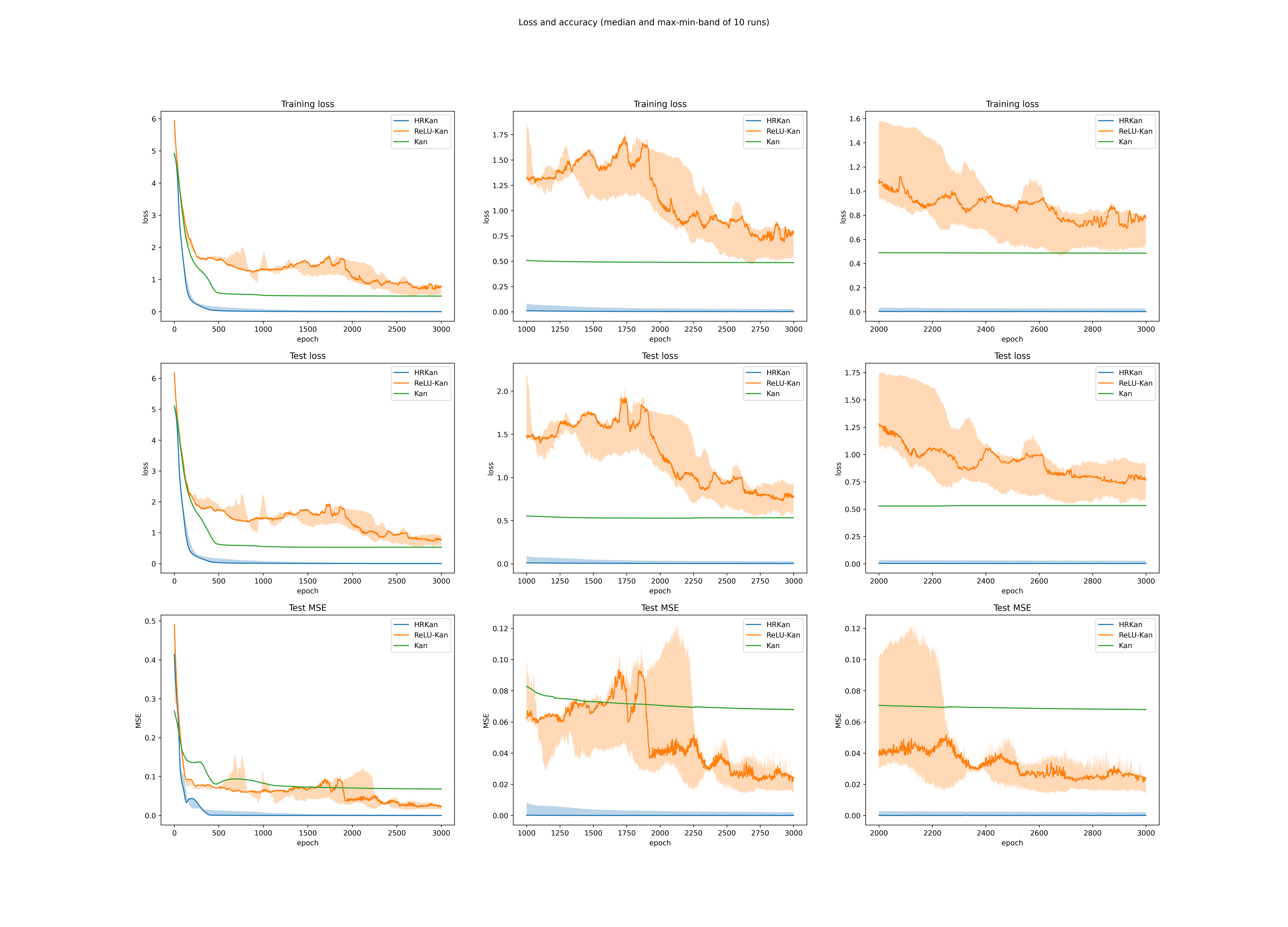}}%
\caption{The median and max-min-band of training loss, test loss and test MSE (MSE between the test-set ground-truth solution and the learnt solutions) of 10 runs for the KAN, ReLU-KAN and HRKAN with the Poisson equation. First column: all 3000 epoches; Second column: last 2000 epoches; Last column: last 1000 epoches.}\label{poisson_loss_band}
\end{figure}

Figure~\ref{poisson_fig} shows the ground-truth solution and the solutions learnt by the KAN, ReLU-KAN and HRKAN and their residual difference compared against the ground-truth solution in one of the 10 runs with the Poisson equation.

Figure~\ref{poisson_loss_band} reveals the median and max-min-band of training loss, test loss and test MSE (Mean square error between the test-set ground-truth solution and the learnt solutions) of 10 runs for the KAN, ReLU-KAN and HRKAN with the Poisson equation. For ease of reading, the first column shows the curves for the entire training process, while the second and last columns show those in the last 2000 and last 1000 epochs respectively.

\begin{table}[t]
\centering
\begin{tabular}[t]{lccc}
 \toprule
   & ReLU-KAN  & HRKAN & KAN\\
 \midrule
 Mean of Test MSE in 10 runs (in $\%$) & 2.18 & \bf{0.0434} & 6.80 \\
 Std. of Test MSE in 10 runs (in $\%$) & 4.28 & \bf{0.129} & 9.43 \\
 Mean of training time in 10 runs (in second) & \bf{21.2} & 27.9 & 109 \\
 \bottomrule
\end{tabular}
\caption{The means and standard deviations of test MSE and means of training time for the KAN, ReLU-KAN and HRKAN in the 10 runs with the Poisson equation.}
\label{poisson_tab}
\end{table}%

Table~\ref{poisson_tab} exhibits the means and standard deviations of test MSE, and means of training time, for the KAN, ReLU-KAN and HRKAN in the 10 runs with the Poisson equation.

There are some key observations to make:
\begin{itemize}
    \item From figures~\ref{poisson_fig}, the last row of figure~\ref{poisson_loss_band} and the first row of table~\ref{poisson_tab}, it can be easily observed that the solutions learnt by HRKANs achieve the highest fitting accuracy (test MSE), while both ReLU-KAN and KAN fail the learn the solution accurately. 
    \item The mean test MSEs attained by ReLU-KANs and KANs are 2.18\% and 6.80\% respectively, whereas the mean test MSE attained by HRKANs is 0.0434\%, at a difference in the scale of $10000\%$. This demonstrates the strong fitting ability of HRKANs.
    \item Also, the standard deviations of test MSEs over the 10 runs are the smallest for HRKANs, at the level of $0.129\%$, whereas those of ReLU-KANs and KANs are $4.28\%$ and $9.43\%$ respectively, at a difference in the scale of $1000\%$. This suggests stronger robustness in the learning of HRKANs.
    \item As shown in figure~\ref{poisson_loss_band}, the training loss, test loss and test MSE of ReLU-KANs keep fluctuating over the training process. One possible reason may be the discontinuity of the higher-order derivatives of activation functions in ReLU-KANs as discussed in section~\ref{s3}. Another possible explanation is the lack of regularization tools in ReLU-KANs compared to KANs, which deserve deeper investigation as potential further study.
    \item The median and max-min-band of training loss, test loss and test MSE of HRKANs and KANs on the other hand show that HRKANs converge relatively rapidly, approaching optimum at around epoch 500, whereas KANs and ReLU-KANs seem to be still converging slowly at the end of the training process (as shown in the 2 plots at the right bottom in figure~\ref{Burgersv_loss_band}). This demonstrates the fast convergence of HRKANs.
    \item Lastly, we can see that the training time of both HRKANs and ReLU-KANs is much smaller than KANs, exhibiting the efficiency of matrix operations in HRKANs and ReLU-KANs. As mentioned before in the problem description, there is possibly extra overhead in computation for KANs because of its steps of grid extension and sparsification, but we may expect such steps won't contribute to most of the computation time for KANs, so our judgement remains valid.
\end{itemize}

\subsection{Burgers' Equation with viscosity}
The Burgers' equation with viscosity is
\begin{equation}
    u_t + u u_x - \nu u_{xx} = 0 
\end{equation}
for $(x,t) \in \left[-5,5\right] \times \left[0,2.5\right]$, with boundary and initial conditions 
\begin{equation*}
  \left\{
    \begin{aligned}
        &u(-5,t) = 0\\
        &u(5,t) = 0\\
        &u(x,0) = \frac{1}{\cosh(x)}
    \end{aligned},
  \right.
\end{equation*}
and $\nu = 0.001$.

We solve for the ground-truth solution using Fast Fourier Transform method~\cite{chen2012parallel}. As we have not included the implementation of grid extension in ReLU-KANs and HRKANs, to make the comparison between KANs, ReLU-KANs and HRKANs fair, we perform a change of variable $y = \frac{x+5}{4}$ to transform the domain into a square domain $[0, 2.5] \times [0, 2.5]$. The resulting PDE is 
\begin{equation}
    u_t + \frac{1}{4} u u_x - \frac{\nu}{16} u_{xx} = 0 
\end{equation}
for $(x,t) \in \left[0,2.5\right] \times \left[0,2.5\right]$, with boundary and initial conditions 
\begin{equation*}
  \left\{
    \begin{aligned}
        &u(0,t) = 0\\
        &u(2.5,t) = 0\\
        &u(x,0) = \frac{1}{\cosh(4x-5)}
    \end{aligned}.
  \right.
\end{equation*}

In the train-set, the number of interior points is $N_{\text{pde}}= 10000$, and the number of boundary points on the 2 boundaries and the number of points for the initial conditions are $N_{\text{bc1}} = 100, N_{\text{bc2}} = 100$ and $N_{\text{ic}} = 100$ respectively. 
The train-set data points on the interior are generated randomly with uniform distribution while those for the boundary and initial conditions are generated uniformly. 
We generate test-set data with a grid with grid size $200\times 200$.

We trained a KAN, a ReLU-KAN and a HRKAN using the loss function
\begin{equation}
L = \alpha L_{\text{pde}} + L_{\text{bc \& ic}}
\end{equation}
where \begin{align*}
    L_{\text{pde}} &= \frac{1}{N_{\text{pde}}}\sum_{i=1}^{N_{\text{pde}}} \left( u_t(x_i,t_i) + \frac{1}{4}u(x_i,t_i) u_x(x_i,t_i) -\frac{\nu}{16} u_{xx}(x_i,t_i) \right)^2,\\
    L_{\text{bc \& ic}} &= \frac{1}{N_{\text{bc1}} + N_{\text{bc2}} + N_{\text{ic}}} \left(\sum_{i=1}^{N_{\text{bc1}}} \left(u(0,t_i)\right)^2 + \sum_{i=1}^{N_{\text{bc2}}} \left(u(2.5,t_i)\right)^2 + \sum_{i=1}^{N_{\text{ic}}} \left(u(x_i,0) - \frac{1}{\cosh(4x_i-5)}\right)^2 \right).
\end{align*}
We set hyperparameter $\alpha = 0.05$.

The structure of the KAN, ReLU-KAN and HRKAN are all of $[2,3,3,3,1]$ with same support for all the bases as $g=7$ and $k=3$. For all three of them, we use the Adam optimizer, set the number of epoch to 3000 and train on the same GPU. We run the same experiment for 10 times. Same as in the previous experiment, we keep the implementation of grid extension and sparsification in KANs unchanged.

\paragraph{Results Discussion.}

\begin{figure}
  \makebox[\textwidth][c]{\includegraphics[width=1.8\textwidth]{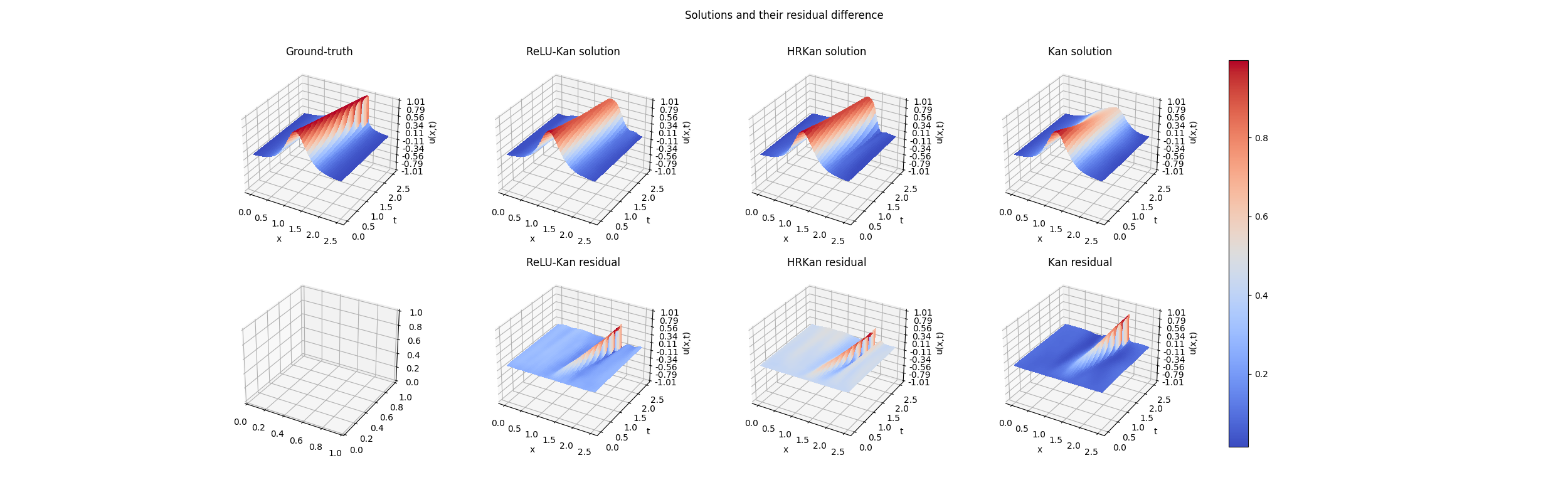}}%
\caption{The ground-truth solution and the solutions learnt by the KAN, ReLU-KAN and HRKAN (Top row) and their residual difference compared against the ground-truth solution (Bottom row) in one of the 10 runs with the Burgers' equation with viscosity.}\label{Burgersv_fig}
\end{figure}

\begin{figure}
  \makebox[\textwidth][c]{\includegraphics[width=1.8\textwidth]{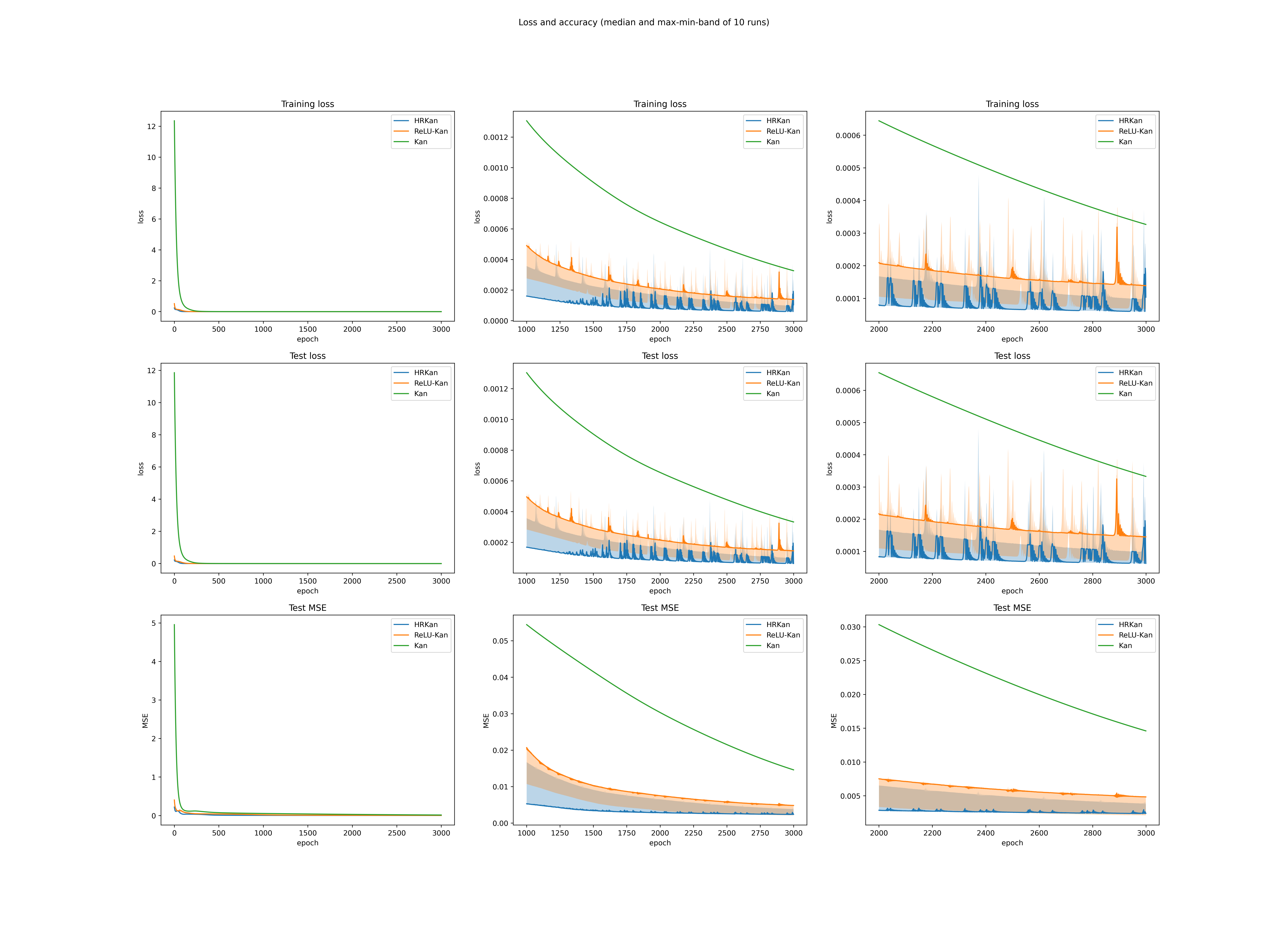}}%
\caption{The median and max-min-band of training loss, test loss and test MSE (MSE between the test-set ground-truth solution and the learnt solutions) of 10 runs for the KAN, ReLU-KAN and HRKAN with the Burgers' equation with viscosity. First column: all 3000 epoches; Second column: last 2000 epoches; Last column: last 1000 epoches.}\label{Burgersv_loss_band}
\end{figure}

Figure~\ref{Burgersv_fig} shows the ground-truth solution and the solutions learnt by the KAN, ReLU-KAN and HRKAN and their residual difference compared against the ground-truth solution in one of the 10 runs with the Burgers' equation with viscosity.

Figure~\ref{Burgersv_loss_band} reveals the median and max-min-band of training loss, test loss and test MSE (Mean square error between the test-set ground-truth solution and the learnt solutions) of 10 runs for the KAN, ReLU-KAN and HRKAN with the Burgers' equation with viscosity. For ease of reading, the first column shows the curves for the entire training process, while the second and last columns show those in the last 2000 and last 1000 epochs respectively.

\begin{table}[t]
\centering
\begin{tabular}[t]{lccc}
 \toprule
   & ReLU-KAN  & HRKAN & KAN\\
 \midrule
 Mean of Test MSE in 10 runs (in $\%$)  & 0.575 & \bf{0.229} & 1.48 \\
 Std. of Test MSE in 10 runs (in $\%$) & 2.23 & \bf{0.861} & 5.85 \\
 Mean of training time in 10 runs (in second) & \bf{61.6} & 75.0 & 624 \\
 \bottomrule
\end{tabular}
\caption{The means and standard deviations of test MSE and means of training time for the KAN, ReLU-KAN and HRKAN in the 10 runs with the Burgers' equation with viscosity.}
\label{Burgersv_tab}
\end{table}%

Table~\ref{Burgersv_tab} exhibits the means and standard deviations of test MSE, and means of training time for the KAN, ReLU-KAN and HRKAN in the 10 runs with the Burgers' equation with viscosity.

There are some key observations to make:
\begin{itemize}
    \item From figures~\ref{Burgersv_fig} and~\ref{Burgersv_loss_band} and table~\ref{Burgersv_tab}, it is obvious that the solutions learnt by HRKANs achieve the highest fitting accuracy (test MSE), while the accuracies of the solutions learnt by ReLU-KANs and KANs are relatively lower.
    
    \item The largest difficulty for all the three models to learn the solution probably lies on the "discontinuity" of the ground-truth solution near $t=2.5$, as displayed in figure~\ref{Burgersv_fig}. Although all three models fail to learn such discontinuity in certain degree, the error in fitting accuracy is the lowest for HRKANs, as shown in figures~\ref{Burgersv_fig}, the last row of figure~\ref{Burgersv_loss_band} and first row of table~\ref{Burgersv_tab}.
    
    \item The mean test MSEs attained by ReLU-KANs and KANs are 0.575\% and 1.48\% respectively, whereas the mean test MSE attained by HRKANs is 0.229\%, demonstrating a stronger fitting ability of HRKANs.

    \item Also, the standard deviations of test MSEs over the 10 runs are the smallest for HRKAN, at the level of $0.861\%$, whereas those of ReLU-KAN and KAN are $2.23\%$ and $5.85\%$. This suggests the stronger robustness in the learning of HRKANs. 

    \item In figure~\ref{Burgersv_loss_band}, the test MSE converges at the fastest speed for HRKAN, approaching the minimum at around epoch 1500, whereas those of ReLU-KAN and KAN are still on the way of slowly converging at the end of the training process. This suggests the fast convergence of HRKANs. The fluctuation of the training and test loss of HRKANs after epoch 1250 may imply over-fitting of HRKANs after epoch 1250 to the "discontinuous" solution. The relatively more fluctuating training loss and test loss in ReLU-KANs and HRKANs may also due to the lack of regularization tools in these two models, compared to KANs.
    
    \item Lastly, we can see that the training time of both HRKAsN and ReLU-KANs is much smaller than KAN, exhibiting the efficiency of matrix operations in HRKANs and ReLU-KANs. As mentioned before, there is possibly extra overhead in computation for KANs because of its steps of grid extension and sparsification, but we expect such steps won't contribute to most of the computation time for KANs, so our judgement remains valid.
\end{itemize}

\section{Conclusion} \label{s5}

In this paper, we proposed a new basis of activation functions for KAN, namely, Higher-order-ReLU (HR). Such Higher-order-ReLU is (1) simpler than the B-Spline basis functions in the original KANs, (2) allows matrix-based operations for fast GPU computing, (3) possesses smooth and non-zero higher-order derivatives essential to Physics-informed neural networks (PINNs). We evaluated Higher-order-ReLU-KANs (HRKANs) on two representative and famous PDEs, namely, a linear Poisson equation, and a non-linear Burgers' equation with viscosity. The detailed experiments exhibit the higher fitting accuracy, stronger robustness and faster convergence of HRKANs. Not only can HRKANs find solutions of PDEs, we expect HRKANs to have strong potential in directions like (1) identifying coefficients of PDEs, (2) finding the optimal control of PDEs, and (3) explaining a neural operator etc. 

\bibliography{mybibfile}

\end{document}